%% file: eccv2022submission.tex
\documentclass[runningheads]{llncs}
\usepackage{graphicx}

\usepackage{tikz}
\usepackage{comment}
\usepackage{amsmath,amssymb} 
\usepackage{color}

\usepackage{graphicx}
\usepackage{amsmath}
\usepackage{amssymb}
\usepackage{booktabs}
\usepackage{times}
\usepackage{epsfig}
\usepackage{graphicx}
\usepackage{kotex}
\usepackage{makecell}
\usepackage{algorithm}
\usepackage{algpseudocode}
\usepackage{caption}
\usepackage{multirow}
\usepackage{multicol}
\usepackage{url}
\usepackage{array}
\usepackage{subcaption}
\usepackage[pagebackref,breaklinks,colorlinks]{hyperref}
\usepackage{breakcites}

\usepackage[accsupp]{axessibility}  

\usepackage{xspace}
\makeatletter
\DeclareRobustCommand\onedot{\futurelet\@let@token\@onedot}
\def\@onedot{\ifx\@let@token.\else.\null\fi\xspace}
\def\eg{\emph{e.g}\onedot} 
\def\ie{\emph{i.e}\onedot}

\makeatother

\usepackage[width=122mm,left=12mm,paperwidth=146mm,height=193mm,top=12mm,paperheight=217mm]{geometry}

\begin{document}
\pagestyle{headings}
\mainmatter
\def\ECCVSubNumber{5099}  

\title{Towards Sequence-Level Training for Visual Tracking} 

\titlerunning{Towards Sequence-Level Training for Visual Tracking}
%
\author{Minji Kim\textsuperscript{1*} \quad
Seungkwan Lee\textsuperscript{3,4*}\quad
Jungseul Ok\textsuperscript{3} \quad
Bohyung Han\textsuperscript{1,2} \quad
Minsu Cho\textsuperscript{3}
}

\authorrunning{M. Kim et al.}
%

\institute{\textsuperscript{1}ECE \& \textsuperscript{2}IPAI, Seoul National University\\
\textsuperscript{3}Pohang University of Science and Technology (POSTECH)\\
\textsuperscript{4}Deeping Source Inc. \\
\url{https://github.com/byminji/SLTtrack}
}

\maketitle
\def\thefootnote{}\footnotetext{* These authors contributed equally to this work.}\def\thefootnote{\arabic{footnote}}

\input{_0_abstract.tex}
\input{_1_introduction.tex}

\input{_2_related_work.tex}

\input{_3_method.tex}
\input{_4_experiment.tex}
\input{_5_conclusion.tex}
\\

\noindent \textbf{Acknowledgments.}
This work was supported by Samsung Advanced Institute of Technology (Neural Processing Research Center), the NRF grants (No. 2021M3E5D2A01023887, No. 2022R1A2C3012210) and the IITP grants (No. 2021-0-01343, No. 2022-0-00959) funded by the Korea government (MSIT).

\clearpage
%
%
\bibliographystyle{splncs04}
\bibliography{egbib}

\input{_6_supple.tex}

\end{document}

%% file: _0_abstract.tex
\begin{abstract}
        
Despite the extensive adoption of machine learning on the task of visual object tracking, recent learning-based approaches have largely overlooked the fact that visual tracking is a sequence-level task in its nature; 
they rely heavily on frame-level training, which inevitably induces inconsistency between training and testing in terms of both data distributions and task objectives. 
This work introduces a sequence-level training strategy for visual tracking based on reinforcement learning and discusses how a sequence-level design of data sampling, learning objectives, and data augmentation can improve the accuracy and robustness of tracking algorithms.
Our experiments on standard benchmarks including LaSOT, TrackingNet, and GOT-10k demonstrate that four representative tracking models, SiamRPN++, SiamAttn, TransT, and TrDiMP, consistently improve by incorporating the proposed methods in training without modifying architectures.
\keywords{visual tracking, sequence-level training, reinforcement learning}

\end{abstract}

%% file: _1_introduction.tex

\section{Introduction}

Visual object tracking aims to estimate the spatial extent, \textit{e.g.}, a bounding box, of a target object over a sequence of video frames~\cite{otb,ex_survey,comp_survey}. 
This task has been drawing significant attention due to its wide range of applications including visual surveillance, robotics, and autonomous driving~\cite{vehicle_tracking,human_tracking}.
Unlike standard recognition tasks such as image classification and object detection, the class of the target object is unknown and only its bounding box at the initial frame is given for testing.
Despite the long history of its study~\cite{ex_survey}, object tracking in the wild still remains challenging due to appearance variation, occlusion, interference, distracting clutter, etc.
To tackle the issues, recent methods increasingly rely on robust feature representations that are learned by deep neural networks with convolutions~\cite{mdnet,rt-mdnet,dimp,siamfc,siamrpn,ocean} and attention~\cite{siamattn,trdimp,transt}.

\begin{figure}[t]
    \begin{center}
        \centering
\begin{minipage}{0.76\linewidth}
    \begin{minipage}{1.0\linewidth}
    \includegraphics[trim={0 0 0 0},clip,width=0.99\linewidth]{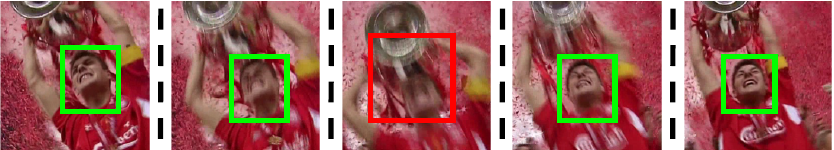}
    \centering\small{(a) Frame-level training} 
    \end{minipage}
    \begin{minipage}{1.0\linewidth}
    \includegraphics[trim={0 0 0 -1.1mm},clip,width=0.99\linewidth]{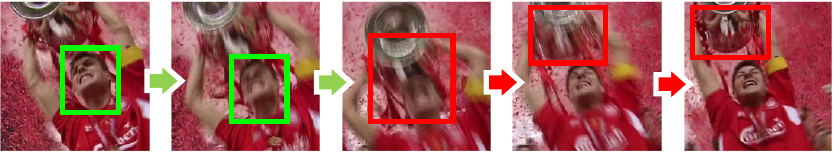}
    \centering\small{(b) Sequence-level self-testing}
    \end{minipage}
\end{minipage} 
\hfill
\begin{minipage}{0.2\linewidth}
    \begin{minipage}{1.0\linewidth}
    \includegraphics[width=1.0\linewidth]{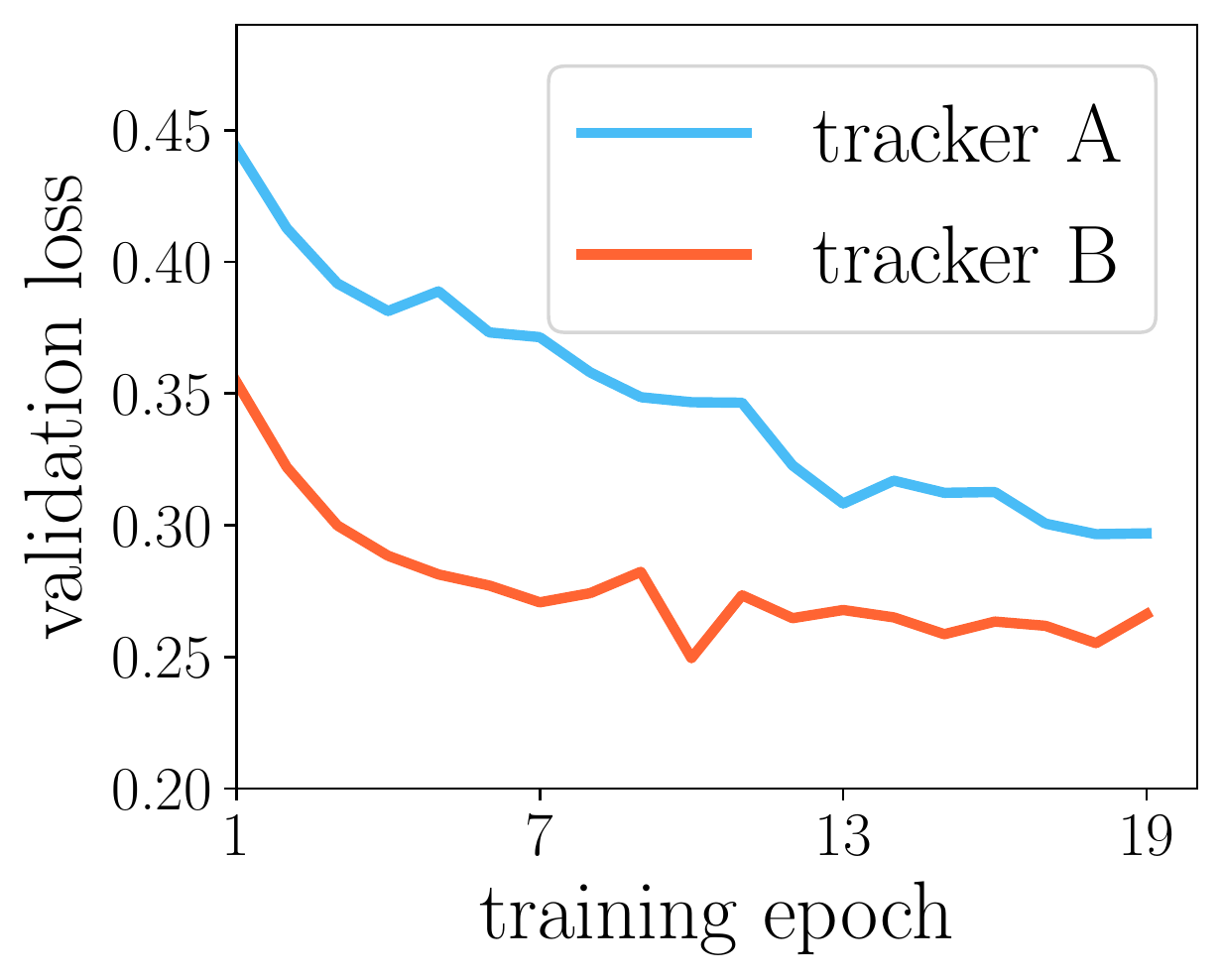}
    \end{minipage}
    \begin{minipage}{1.0\linewidth}
    \includegraphics[width=1.0\linewidth]{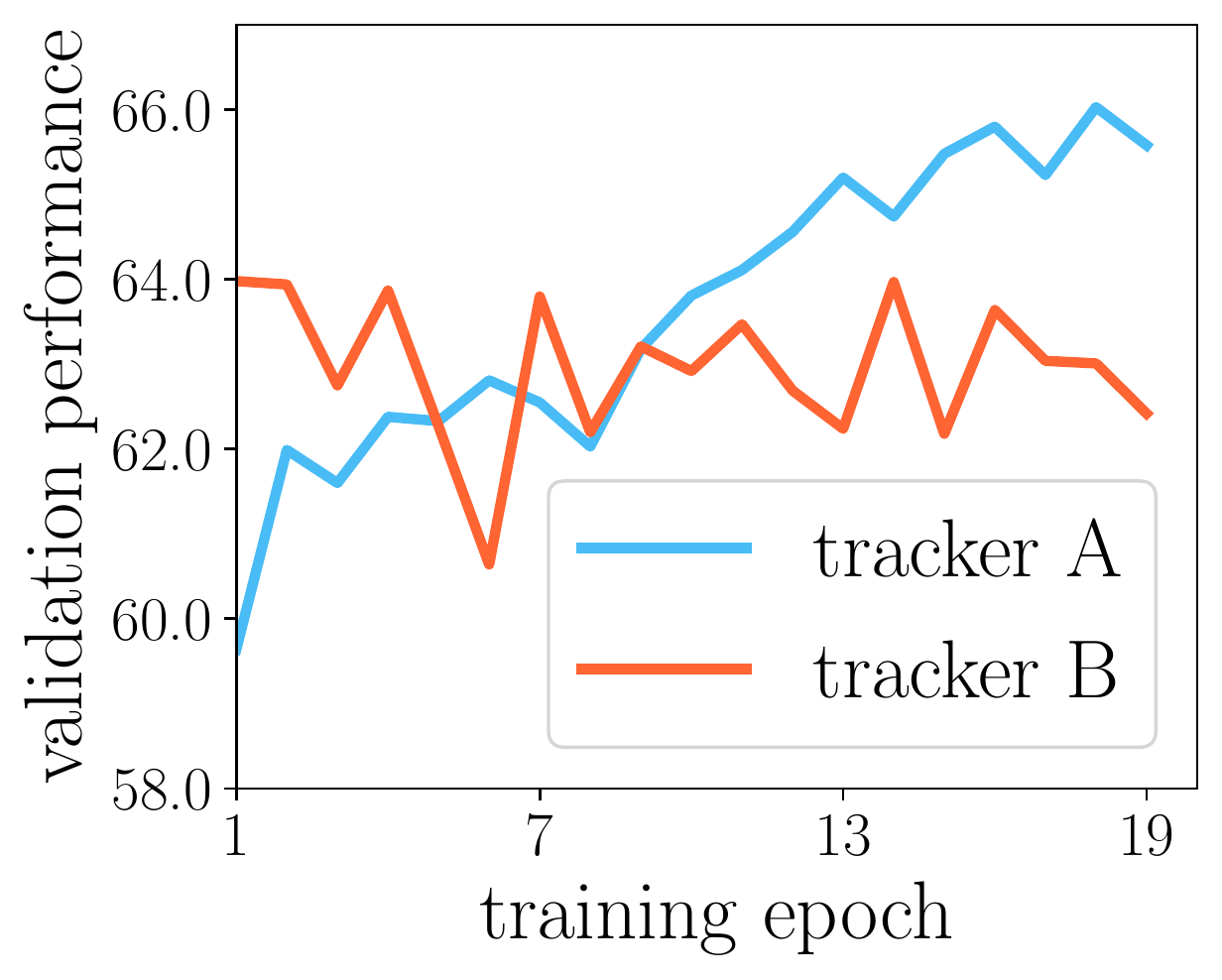}
    \centering\small{(c) Loss \textit{vs.} Perf.}
    \end{minipage}
\end{minipage}
        \captionof{figure}
        {
          Pitfall of frame-level training for visual tracking. Training a tracker to better localize a target in each of the individual frames of (a) does not necessarily improve actual tracking in the sequence of (b). Green/red boxes indicate success/failure in localization. Due to the issue, inconsistency between the loss and the performance is often observed during training as shown in (c) where trackers, A and B, being frame-level trained are evaluated while the validation loss (top) and the tracking performance of average overlap (bottom) are measured. After 10 epochs, A outperforms B in spite of higher losses.
        }\label{fig:teaser}
    \end{center}%
\end{figure}

Although learning to track has been widely adopted in the research community, it is largely overlooked that visual tracking is essentially a {\em sequence-level} task; the estimated target state in the current frame is affected by the history of target states in the previous frames and also influences tracking results in the subsequent frames. 
For example, recent state-of-the-art methods~\cite{mdnet,dimp,siamban,prdimp,siamrpnpp,siamattn,trdimp,ocean,alpharefine} rely heavily on {\em frame-level} training, which encourages the trackers to better localize target objects in each frame through supervised learning. 
While it greatly improves the tracker by learning robust features for tracking, disregarding the sequential dependency across frames can lead to unexpected tracking failures.   
Let us assume a tracker that is trained with a typical frame-level training scheme using a set of annotated training videos; 
random pairs of a target template and a search frame are sampled from a video and the tracker is trained to best localize the target on the search frame independently for each pair. 
As shown in Fig.~\ref{fig:teaser}, now consider one of the training videos that contains a {\em hard} frame, where the tracker fails to localize the target object.
Although the tracker is trained to perform almost perfectly on frame-level localization except for the hard frame (Fig.~\ref{fig:teaser}a), its sequence-level performance may turn out to be poor as it loses the target from the hard frame in actual tracking on the sequence (Fig.~\ref{fig:teaser}b). 

This pitfall of frame-level training mainly stems from inconsistency between training and testing in terms of both {\em data distributions} and {\em task objectives}. 
First, the tracker observes data samples, \ie, tracking situations, that significantly deviate from a real data distribution.
That is, while in actual tracking the search window at each frame is determined based on the estimation at the previous frame, in the frame-level training it is not; the search window is typically sampled by adding a random transformation to the ground-truth bounding box.
Second, the tracker learns with an objective, \ie, a reward system, that is largely different from actual tracking. The tracking performance in testing puts significant importance on retaining localization accuracy over a sequence, whereas it is only immediate localization quality that matters in the frame-level training.  
The mismatch of the objectives between training and testing often leads to unexpected results as shown in Fig.~\ref{fig:teaser}c; two trackers, A and B, being trained with the same network architecture and the same frame-level objective, are tested on the GOT-10k validation split where the loss and the performance are measured\footnote{Both trackers A and B adopt SiamRPN++ as their network architectures, but the backbone of tracker A is frozen in the early training stages.}.
While constantly yielding a higher loss, tracker A achieves better performance than tracker B after 10 epochs. Such inconsistency should be rectified for more robust tracking but has hardly been explored so far in the tracking community. 

This work investigates the sequence-level training for visual object tracking and analyzes how the performance of a tracking algorithm improves by resolving the aforementioned inconsistency issues.
Without adding any architectural components, we train a tracker end-to-end by simulating sequence-level tracking scenarios with a properly matching reward system in the framework of reinforcement learning (RL).
Specifically, our tracker observes a sequence of frames sampled from an actual tracking trajectory and optimizes the objective based on the test-time metric such as the average overlap~\cite{otb,got10k}.
Our training strategy not only resolves the inconsistency in data distributions but also addresses the discrepancy in task objectives by teaching the tracker how its decision in the current frame affects future ones.
Furthermore, this approach enables us to extend data augmentation to a temporal domain; on top of commonly-used data augmentation strategies in the spatial domain~\cite{comp_survey}, we can simulate temporally-varying tracking scenarios for training, corresponding to videos with diverse object/camera motions~(Sec.~\ref{sec:method}). 
Note that this new type of augmentation has not been available under the frame-level training of previous trackers. 
To sum up, the proposed sequence-level training allows a tracker to learn a robust strategy for realistic tracking scenarios by leveraging the simulated tracking samples, (\ie, {\em sequence-level sampling}), the long-term objective (\ie, {\em sequence-level objective}), and the data augmentation in the temporal domain (\ie, {\em sequence-level augmentation}).

Our contributions are summarized as follows.
\begin{itemize}
    \item[$\bullet$] We analyze the inherent drawbacks of frame-level training adopted in recent trackers,
    which motivate sequence-level training (SLT) for robust visual object tracking.
    \item[$\bullet$] We introduce an SLT strategy for visual tracking in an RL framework and propose an effective toolset of data sampling, training objectives, and data augmentation. 
    \item[$\bullet$] We demonstrate the effectiveness of SLT using four recent trackers, SiamRPN++~\cite{siamrpnpp}, SiamAttn~\cite{siamattn}, TransT~\cite{transt}, and TrDiMP~\cite{trdimp}, and achieve competitive performance on the standard benchmarks, LaSOT~\cite{lasot}, TrackingNet~\cite{trackingnet}, and GOT-10k~\cite{got10k}.
    \item[$\bullet$] We provide in-depth analyses of SLT by studying the effects of the sequence-level data sampling and the corresponding objective as well as the sequence-level data augmentation diversifying tracking episodes in a temporal domain.
\end{itemize}

%% file: _2_related_work.tex

\section{Related Work}
\subsection{Visual Object Tracking (VOT)}

There has been a large body of active research on VOT~\cite{ex_survey,comp_survey,dcf_siam_survey}, which still remains one of the major topics in computer vision. Recent methods for VOT have greatly improved tracking performance based on deep learning with large-scale datasets~\cite{yt-bb,imagenet,coco,trackingnet,got10k,lasot}.
Currently, state-of-the-art trackers are represented by two families of trackers: Siamese~\cite{siamfc,siamrpn,siamrpnpp,siamattn,transt} and DiMP~\cite{atom,dimp,prdimp,trdimp}. 
The Siamese trackers~\cite{siamfc,siamrpn,siamrpnpp} rely heavily on an effective template-matching mechanism that is trained offline using large-scale datasets. While being fast and accurate in a short term, they tend to be vulnerable to long-term tracking due to the lack of online adaptability. 
Recent variants mitigate this limitation by updating template features during tracking~\cite{updatenet,dasiamrpn} or using an attention mechanism to diversify the feature representation~\cite{siamattn}.
DiMP trackers~\cite{dimp,prdimp,trdimp} learn the online model predictor for target center regression and combine it with bounding box regression.
Their model predictor, which is an iterative optimization-based neural module, is trained offline with a meta-learning-based objective and is used to build an online target model during tracking.
Recently, Transformer-based architectures are adopted in both Siamese~\cite{transt} and DiMP~\cite{trdimp} trackers to enhance target feature representations.

Note that all of these state-of-the-art trackers are trained with frame-level objectives; 
it forces the model to take an instantaneous greedy decision at each frame, ignoring that the tracking errors accumulate over the sequence.
In contrast, we study how to improve tracking by considering temporal dependency in the sequence of frames.

\subsection{Reinforcement Learning for VOT}
Our sequence-level training scheme is built on reinforcement learning (RL), which provides a natural framework for sequential decision-making on the problem with interactive temporal dependency.
RL is not new in visual tracking and there exist several RL-based trackers~\cite{hp-siam,pomdp,east,drl-is,adnet,act,dtnet,pacnet}.
Most of them~\cite{hp-siam,drl-is,pomdp,east} aim to assist tracking by learning an additional RL agent while considering both a given tracker and its input sequence as an environment. 
HP-Siam~\cite{hp-siam} uses RL to optimize hyper-parameters of the tracker such as scale step, penalty, and window weight.
DRL-IS~\cite{drl-is} and P-Track~\cite{pomdp} learn a policy to decide the state transition of the tracker.
EAST~\cite{east} uses RL to speed up tracking by learning to stop feed-forwarding frames through layers.

Only a few RL-based methods~\cite{adnet,act,pacnet} learn the tracker itself as an RL agent, which performs actual tracking, \eg, estimating a target bounding box.
ADNet~\cite{adnet} formulates tracking as a discrete box adjustment problem, where at each time step the agent observes a current frame and a previous localization box and then decides discrete actions for adjusting the box. In a similar manner, ACT~\cite{act} learns to predict box transformation parameters in a continuous space of actions while PACNet~\cite{pacnet} jointly learns both box estimation and state transition of the tracker.
Their methods, however, require a specific form of output for training trackers in RL, \eg, the output of box transformation parameters~\cite{act,pacnet} and pre-defined actions for box adjustment~\cite{adnet}, and hardly exploit the advantage of RL training in a generic perspective.
In contrast, we introduce a generic training strategy using RL and analyze its advantages over its frame-level counterpart, which is prevalent in recent state-of-the-art trackers. 
We advocate sequence-level training {\em per se} as the integral role of RL in tracking, showing that recent learnable trackers greatly benefit from RL-based training without additional components.  

Regarding our effort to address the training-testing inconsistency, the most related work is \cite{scst}, which tackles a similar problem in image captioning. 
It proposes self-critical sequence training (SCST), a form of the well-known REINFORCE~\cite{reinforce} algorithm, to train image captioning models directly on NLP metrics.  
We build our sequence-level training scheme on SCST and adapt it for visual object tracking.

%% file: _3_method.tex

\section{Our Approach}\label{sec:method}

\subsection{Sequence-Level Training (SLT)}
\label{sec:RL}

Given a video $\mathbf{v} = (v_0, ..., v_T)$ of $T+1$ frames and the ground-truth bounding box $g_0$ of the target object in the initial frame $v_0$, a tracker sequentially predicts a bounding box $l_t$ of the target in each frame $v_t$, $t =1, 2, ..., T$.
The tracker, parameterized by $\theta$, is modeled as a function $\pi_\theta$ that
takes observation $o_t$ and predicts $l_t$, \ie, $l_t = \pi_\theta(o_t)$, where $o_t$ is the information available at time $t$ including the video frames $(v_0, ..., v_t)$, the initial target bounding box $g_0$, and the previously estimated target states $(l_1, ..., l_{t-1})$.
In online tracking, most trackers estimate the current state based on the previous prediction $l_{t-1}$ and the observation of the current frame $v_{t}$, \eg, searching for the optimal local window in $v_{t}$ around $l_{t-1}$.
The objective of a tracking algorithm is to maximize the \textit{sequence-level} performance
$r(\mathbf{l})$, where $\mathbf{l} = (l_1, ..., l_T)$ is the estimated target states in a video and $r$ is an evaluation metric, \eg, average overlap ratio~\cite{otb,got10k}.

Due to the sequential structure of the tracking process, its current decision $l_t$ naturally affects the future ones $l_{t+1}, ..., l_T$.
In the frame-level training, which is the {\em de facto} standard for recent methods~\cite{siamrpnpp,siamattn,transt,dimp,trdimp,stark}, however, trackers do not simulate sequential target state estimation procedure in training.
In other words, given each frame $v_t$, trackers are trained to localize the target object bounding box $g_t$ from its random perturbation $\rho(g_t)$ instead of $l_{t-1}$, where $\rho$ is a random perturbation function.
Such a frame-level approximation, \ie, $l_{t-1} \approx \rho(g_t)$, requires additional hyper-parameters for $\rho$ and, more importantly, introduces inconsistency of data distributions between training and testing; the trackers have no opportunity to learn how the previous decision affects the current one in a real tracking scenario. 

To overcome the limitation of frame-level training and capture the temporal dependency between decisions, we build our sequence-level training scheme based on RL.
We simulate the tracker (agent) on a sequence of video frames, and directly optimize it with respect to the test-time evaluation metric:
\begin{equation}
\label{eq:obj-RL}
L(\theta) := - \mathbb{E}_{\mathbf{l} \sim \pi_\theta} [r(\mathbf{l})] \;.
\end{equation}
Note that this \textit{sequence-level objective} directly optimizes the real objective of tracking, and thus is a more natural way to train trackers than frame-level counterparts.
This objective relieves the aforementioned issues in frame-level training; the tracker observes the real data distributions via \textit{sequence-level sampling}, which draws samples from actual tracking trajectories and facilitates learning temporal dependency in tracking.

To directly optimize the task objective in~\eqref{eq:obj-RL}, we employ the REINFORCE algorithm~\cite{reinforce}. 
According to the algorithm, the expected gradient is computed as follows:%
\begin{equation}
\begin{split}
&\nabla_\theta L(\theta) = - \mathbb{E}_{\mathbf{l} \sim \pi_\theta} [r(\mathbf{l}) \nabla_\theta \log p_\theta(\mathbf{l})].
\end{split}
\end{equation}
In practice, the expected gradient is approximated by using a single Monte Carlo sample $\mathbf{l} = (l_1, ..., l_T)$ of sequential decisions from $\pi_\theta$. 
For each training episode, the gradient is given by
\begin{equation}
\begin{split}
&\nabla_\theta L(\theta) \approx - r(\mathbf{l}) \nabla_\theta \log p_\theta(\mathbf{l}).\\
\end{split}
\end{equation}

\begin{figure*}[t]
\centering
\includegraphics[trim={0 0 7mm 0},clip, width=\linewidth]{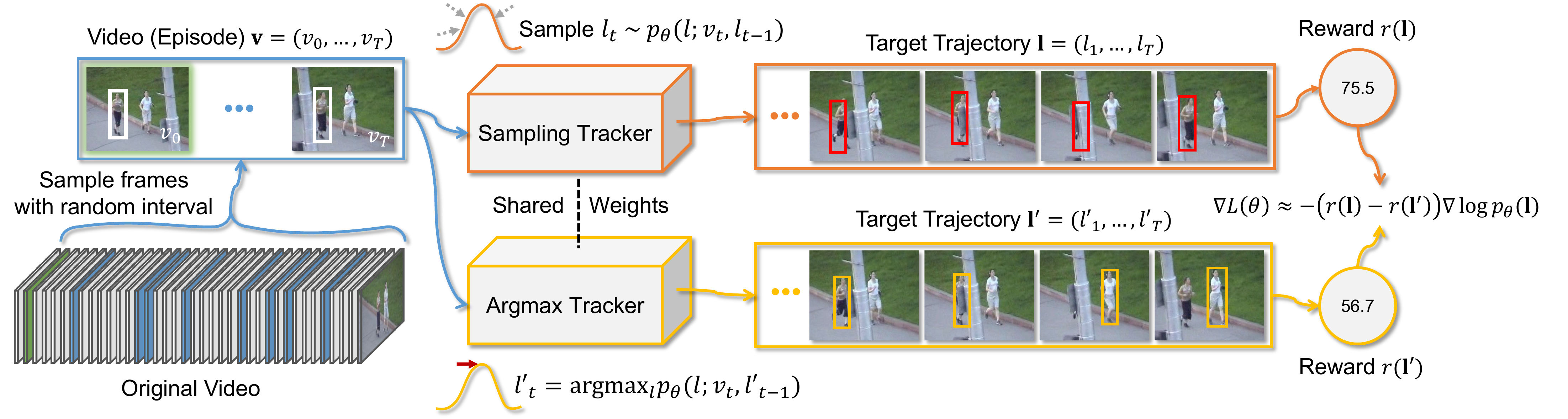}
\caption{
Illustration of our sequence-level training framework.
In training time, a video (episode), which is sampled from the original video with random intervals, is tracked twice by the sampling tracker and the argmax tracker.
In this example, when a target person is fully occluded for a while, the argmax tracker mistakenly localizes the other person as the target due to its highest score in the occluded scene. In contrast, the sampling tracker stays nearby the previously estimated location (because of the random sampling) and successfully re-tracks the target object. In such a case, the reward becomes positive so that the sampled action is encouraged. In the opposite case, the reward becomes negative so that the sampled action is discouraged.
}
\label{fig:scst}
\end{figure*}

To reduce the variance of gradient estimation, we adopt the self-critical sequence training (SCST)~\cite{scst}, which exploits the test-mode performance of the current model as a baseline for the reward.
To be specific, we adopt two trackers sharing network parameters: a sampling tracker and an argmax tracker.
During training, a video (episode) is played twice independently by both trackers.
Given the probability distribution of actions, the sampling tracker decides the target bounding box stochastically while the argmax tracker selects the most confident one.
If the agent obtains a higher reward from the sampling mode than the argmax mode, the resulting reward becomes positive to encourage the sampled actions.
Otherwise, the agent receives a negative reward to suppress the sampled actions.
In this way, we employ the SCST algorithm to train the tracker using the following gradient:
\begin{equation}
\begin{split}
&\nabla_\theta L(\theta) \approx - (r(\mathbf{l}) - r(\mathbf{l}^\prime)) \nabla_\theta \log p_\theta(\mathbf{l}),\\
\end{split}
\label{scst_gradient}
\end{equation}
where $r(\mathbf{l})$ and $r(\mathbf{l}^\prime)$ are rewards obtained from the current model by the sampling mode and the argmax mode during training, respectively. 
Such a REINFORCE-based training scheme is a certain realization of SLT and may be further improved by other RL-based algorithms.
We illustrate the proposed sequence-level training pipeline in Fig.~\ref{fig:scst}, and provide the pseudo-code in Algorithm~\ref{training_algorithm}.

\begin{algorithm}[t]
	\caption{Sequence-Level Training}
	\label{training_algorithm}
	\begin{algorithmic}[1]
		\Procedure{Sequence-level Training}{}\newline
		\textbf{Input:} A tracker parametrized by $\theta$, training dataset $\Gamma$
		\While{not converged}
		    \State Sample a video $\mathbf{v} = (v_0, ..., v_T)$ and ground-truth $\mathbf{g} = (g_0, ..., g_T)$ from $\Gamma$
		    \State Initialize the tracker by using $\{v_0, \ g_0\}$
		    \State $l_0 = g_0$ \Comment{Initial target location for the sampling tracker}
		    \State $l_0' = g_0$ \Comment{Initial target location for the argmax tracker}
    		\For{$t \in {1,...,T}$}
    		    \State $l_t = $ sample $l$ from $p_\theta(l ; v_t, l_{t-1})$
    		    \State $l'_t = \arg\max_l {p_\theta(l ; v_t, l'_{t-1})}$
    		\EndFor
    		\State $r = \text{evaluate}(\{l_1, ..., l_T\}, \{g_1, ..., g_T\})$ 
    		\State $r' = \text{evaluate}(\{l'_1, ..., l'_T\}, \{g_1, ..., g_T\})$ 
    		\State $L = -(r - r') \sum_{t=1}^{t=T} \log p_\theta (l_t; v_t, l_{t-1})$ \Comment{(\ref{scst_gradient})}
    		\State $\theta = \theta - \alpha \nabla_\theta L$ \Comment{Update model parameters}
		\EndWhile
		\EndProcedure
	\end{algorithmic}
\end{algorithm}

\subsection{Integration into Tracking Algorithms}
\label{sec:integration}
We now present how to integrate the proposed SLT scheme into existing trackers.
To demonstrate the effect of SLT, we adopt four representative trackers as our baselines: SiamRPN++~\cite{siamrpnpp}, SiamAttn~\cite{siamattn}, TransT~\cite{transt}, and TrDiMP~\cite{trdimp}. In the following, we briefly describe each tracker and explain how it is trained.

\subsubsection{SLT-SiamRPN++.}
SiamRPN++~\cite{siamrpnpp} is a representative Siamese tracker based on the region proposal network (RPN)~\cite{fast-rcnn}.
This method tracks the target object by repeatedly matching between feature embeddings of a template image and a search image~\cite{siamfc}.
Specifically, the tracker outputs confidence scores and box coordinates of $N$ anchor boxes and performs greedy selection to choose the most confident one, where its box coordinates are selected as the estimation of the target location in the current frame.
Since the current prediction for the target state, \ie, position and size, determines the search area in the next frame, this decision not only influences the current frame but also will potentially affect predictions in the future.
Thus, we reinforce the box selection procedure of SiamRPN++ with the proposed sequence-level training strategy to teach the tracker temporal dependencies.

Since our training method assumes that the target localization of the tracker is a stochastic action, we convert the greedy anchor selection of SiamRPN++ to become stochastic.
Let $\mathbf{x} = (x_1, ..., x_{N}) \in \mathbb{R}^{N}$ denote the output anchor scores of SiamRPN++, where $N = a \times H \times W$ denotes the number of candidates in a score map with size of $H \times W$ and $a$ anchor types.
We define a categorical distribution $p(n)$ as follows:
\begin{equation}
\begin{split}
& p(n) = \frac{\exp(\sigma^{-1}(x_n))} {\sum_{m=1}^{N} \exp(\sigma^{-1}(x_m))}, \\
\end{split}
\label{eq:sampling_prob}
\end{equation}
where $\sigma^{-1}$ indicates the logit (inverse sigmoid) function.
Since $x_n$ is a normalized score whose value is between 0 and 1, we first apply the logit function before applying the softmax function.
In training time, the tracker samples an anchor box from $p$ for the current target localization.
In test time, it selects the most confident anchor box deterministically as in the original SiamRPN++.
For each training episode, the loss in the classification branch is given by
\begin{equation}
\begin{split}
L = -(r(\mathbf{l}) - r(\mathbf{l}^\prime)) \sum_{t=1}^{T} \log p(n_t), \\
\end{split}
\label{eq:slt_loss_siamrpnpp}
\end{equation}
where $(r(\mathbf{l}) - r(\mathbf{l}'))$ is the self-critical reward (Sec. 3.1) and $n_t$ is the sampled anchor box at frame $t$.
The overall sequence-level training loss of SiamRPN++ is defined by combining the loss $L$ for the classification branch and the $\ell_1$ loss for the bounding-box regression branch~\cite{siamrpnpp}, which is given by
\begin{equation}
\begin{split}
L_{\textrm{siamrpn++}} = L + L_{\textrm{bbox}}. \\
\end{split}
\label{eq:objective_siamrpnpp}
\end{equation}

\subsubsection{SLT-SiamAttn.}
SiamAttn~\cite{siamattn} is an extension of SiamRPN++ with an additional bounding box refinement module and a mask prediction module, along with attention modules for enhancing the feature representation.
Similar to SiamRPN++, SiamAttn takes greedy anchor selection to choose the best target candidate among $N$ anchor boxes from RPN.
Once the box is chosen, SiamAttn additionally refines the bounding box using a deformable RoI pooling operation~\cite{deformableconv}.
We thus use the same loss $L$ for the classification branch.
The overall sequence-level training loss of SiamAttn is to enhance the capability of target classification with SLT and increase the accuracy of localization modules with the help of sequence-level data sampling: 
\begin{equation}
\begin{split}
L_{\textrm{siamattn}} = L + \lambda_1 L_{\textrm{bbox}}+ \lambda_2 L_{\textrm{refine-bbox}}+ \lambda_3 L_{\textrm{mask}}, \\
\end{split}
\label{eq:objective_siamattn}
\end{equation}
where each loss term except for $L$ follows~\cite{siamattn}. The weight parameters are set as $\lambda_1 = 0.5$, $\lambda_2 = 0.5$, and $\lambda_3 = 0.2$.

\subsubsection{SLT-TransT.}
TransT~\cite{transt} adopts Transformer-like feature fusion networks into Siamese architecture and localizes the target object by computing attention between template vectors and search vectors.
Unlike SiamRPN++ and SiamAttn, TransT has neither anchor points nor anchor boxes, and its prediction heads directly make $N$ classification results and $N$ normalized box estimations from fusion vectors corresponding to each position of the feature map, where $N = H \times W$ denotes the size of the feature map. 
We also take a categorical distribution $p(n)$ of equation (\ref{eq:sampling_prob}) for $N$ candidate vectors, and use the same loss $L$ for the classification branch.
The overall sequence-level training loss for TransT is:
\begin{equation}
\begin{split}
L_{\textrm{transt}} &= L + \lambda_4 L_{\textrm{bbox-L1}} + \lambda_5 L_{\textrm{bbox-GIoU}}, \\
\end{split}
\label{eq:objective_transt}
\end{equation}
where the box regression losses follow~\cite{transt} and $\lambda_4 = 0.33$, $\lambda_5 = 0.13$
in our implementation.

\subsubsection{SLT-TrDiMP.}
TrDiMP~\cite{trdimp} is one of the state-of-the-art DiMP tracker using Transformer architectures.
Its tracking procedure consists of two steps: target center prediction and bounding box regression.
Given template samples generated from the initial frame, the model predictor generates a discriminative CNN kernel to convolve with the feature embedding from a search image for target response generation.
The most confident location of the score map becomes the target center prediction, and starting from the randomly drawn candidate boxes around the location, the final bounding box estimation $l_t$ is obtained from the IoU-Net-based box optimization~\cite{atom,prdimp}.

For sequence-level training of TrDiMP, we convert the target center prediction into stochastic action by simply taking softmax on its score prediction.
Let $\mathbf{y} = (y_1, ..., y_{N}) \in \mathbb{R}^{N}$ denotes the center prediction score of TrDiMP, where $N = H \times W$ means a number of candidates in a score map with a spatial size of $H \times W$.
Now we define a categorical distribution $p(n)$ as follows:
\begin{equation}
\begin{split}
& p(n) = \frac{\exp(y_n)} {\sum_{m=1}^{N} \exp(y_m)}. \\
\end{split}
\label{eq:sampling_prob_trdimp}
\end{equation}
The loss for the center prediction module is same with $L$, and the overall sequence-level training loss for TrDiMP is:
\begin{equation}
\begin{split}
L_{\textrm{trdimp}} &= L + \lambda_6 L_{\textrm{iou-net}}, \\
\end{split}
\label{eq:objective_trdimp}
\end{equation}
where the loss term for IoU-Net follows~\cite{trdimp} and $\lambda_6$ is set to 0.0025.

\subsection{Sequence-Level Data Augmentation}
Learning visual tracking in a sequence level naturally motivates sequence-level data augmentation that is conceptually incompatible with frame-level training.
To improve the data quality and avoid the over-fitting problem of the networks, data augmentation strategies in a spatial domain such as geometric transformation, color perturbations, and blur, are widely adopted for convolutional trackers~\cite{comp_survey}.
Conventional frame-level training, however, treats the training sequence merely as a group of independent images that need to be sampled and cropped, hardly considering the relationship between each image, thereby missing the potential effect of data augmentation towards the temporal domain.
In contrast, our sequence-level training effectively benefits from exploring diverse changes in the temporal axis that can enrich the tracking scenarios, as well as the conventional data augmentation strategies in the spatial axis.

Among many possible ways of sequence-level augmentation, here we focus on a simple frame-interval augmentation, \ie, subsampling the training videos with different frame intervals.
In our sequence-level augmentation setting, some episodes are sampled from the original video with random intervals as shown on the left side of Fig.~\ref{fig:scst}.
This scheme simulates dynamic visual differences along time steps and teaches the tracker to adapt to situations in which objects and/or cameras move faster.
Diversifying the tracking scenarios in terms of frame rates makes the tracker improve or at least maintain the performance in general test videos.
Experimental results in Sec.~\ref{sec:experiments} show how our augmentation strategy affects the tracking performance.
We believe that more advanced sequence-level augmentation strategies, \eg, temporal motion blur, may help sequence-level training further in general.

%% file: _4_experiment.tex
    
\section{Experiments}\label{sec:experiments}
This section presents the effectiveness of the proposed sequencel-level training using four baseline trackers, SiamRPN++~\cite{siamrpnpp}, SiamAttn~\cite{siamattn}, TransT~\cite{transt}, and TrDiMP~\cite{trdimp}, 
on three standard benchmarks, LaSOT~\cite{lasot}, TrackingNet~\cite{trackingnet}, and GOT-10k~\cite{got10k}.

\subsection{Implementation Details}
As widely adopted in deep RL~\cite{deepqlearning,scst,adnet}, we pre-train the trackers with supervised learning, which is done with frame-level training in our case, to stabilize and speed up the subsequent SLT. 
For each training iteration, $k$ tracking episodes are randomly sampled from training datasets, where $k$ is set to 8 for SiamRPN++, TransT, and TrDiMP and 12 for SiamAttn, respectively.
Each episode is composed of $T$ video frames and a single template frame and $T$ is a hyper-parameter for training.
For frame-interval augmentation, the interval is randomly chosen every time sampling the video frames and its maximum is set to 7 for SiamRPN++ and SiamAttn, and 10 for TransT and TrDiMP, respectively.
We use the average overlap (AO) score for the reward function $r$ in Eq.~\ref{eq:slt_loss_siamrpnpp}.

Many recent trackers have post-processing strategies based on geometric priors~\cite{siamrpnpp,siamattn,dimp,trdimp,transt}.
For example, SiamRPN++ has the cosine-window penalty and the shape penalty, which prevent drastic updates in target bounding box estimation.
These penalties are typically not applied during frame-level training, which also brings inconsistency between training and testing.
However, our sequence-level training also resolves this inconsistency problem. 
Note that $x_n$ in Eq.~\ref{eq:sampling_prob} is the anchor score after post-processing.

The argmax and sampling trackers in our SLT framework share all weights for training, and our tracker behaves like the argmax tracker during inference.
Thus, the additional memory cost is marginal in training, and the test-time efficiency of the original tracker is not affected by SLT.
Our algorithm is implemented in Python using PyTorch with NVIDIA RTX A6000 2GPUs.

\subsection{Training Dataset}

For a fair comparison, we aligned the pre-training datasets for the baseline with the fine-tuning datasets for SLT as similar as possible.
1) For SiamRPN++, we adopt LaSOT, TrackingNet, and GOT-10k for both pre-training and fine-tuning.
2) Following the original paper, SiamAttn is trained on LaSOT, TrackingNet, COCO~\cite{coco}, and YouTube-VOS~\cite{ytvos}.
Since COCO is an image dataset, a data augmentation scheme such as shift, scale, and blur is adopted to extend the image to compose an episode.
Note that the data augmentation strategy except for frame-interval augmentation is used only for COCO.
3) Finally, TransT and TrDiMP are both pre-trained using  LaSOT, TrackingNet, GOT-10k, and COCO, as same as the original papers, and then fine-tuned on three video datasets, LaSOT, TrackingNet, and GOT-10k.

\subsection{Evaluation}

We compare the performance of SLT with four baseline trackers. Note that for a fair comparison, we strictly maintain the same test-time hyper-parameters for each method for all datasets.
Only TrDiMP uses a different hyper-parameter setting for evaluation in LaSOT following the baseline paper~\cite{trdimp}.

\begin{table*}[t]
\centering
\caption{
Performance of sequence-level training on LaSOT, TrackingNet, and GOT-10k.
}
\label{table:eval_base_slt}
\scalebox{0.83}{
\setlength{\tabcolsep}{6pt}
\begin{tabular}{cc|cc|ccc|ccc}
\Xhline{2\arrayrulewidth}
\multicolumn{2}{c|}{\multirow{2}{*}{Method}} & \multicolumn{2}{c|}{LaSOT} & \multicolumn{3}{c|}{TrackingNet} & \multicolumn{3}{c}{GOT-10k}\\  
   &        & AUC ($\Delta$) & $\text{P}_{\text{Norm}}$ & AUC ($\Delta$) & $\text{P}_{\text{Norm}}$ & P & AO ($\Delta$) &$\text{SR}_{0.5}$ &$\text{SR}_{0.75}$ \\ \hline
\multirow{2}{*}{SiamRPN++} & Base & 51.0 & 60.3 & 68.2 & 78.3 & 68.9  & 49.5 & 58.0 & 30.5\\
                           & +SLT & 58.4 (+7.4) & 66.6 & 75.8 (+7.6) & 81.0 & 71.3 & 62.1 (+12.6) & 74.9 & 49.0\\ 
                          \hline
\multirow{2}{*}{SiamAttn}  & Base  & 54.8 & 63.5 & 74.3 & 80.9 & 70.6 & 53.4 & 61.8 & 36.4\\
                           & +SLT & 57.4 (+2.6) & 66.2 & 76.9 (+2.6) & 82.3 & 72.6 & 62.5 (+9.1) & 75.4 & 50.2\\ 
                          \hline
\multirow{2}{*}{TrDiMP}    & Base & 63.3 & 72.3 & 78.1 & 83.3 & 73.1 & 67.1 & 77.4 & 58.5\\
                           & +SLT & 64.4 (+1.1) & 73.5 & 78.1 (+0.0) & 83.1 & 73.1 & 67.5 (+0.4) & 78.8 & 58.7 \\ 
                          \hline
\multirow{2}{*}{TransT}    & Base & 64.2 & 73.7 & 81.1 & 86.8 & 80.1 & 66.2 & 75.5 & 58.7\\
                           & +SLT & 66.8 (+2.6) & 75.5 & 82.8 (+1.7) & 87.5 & 81.4 & 67.5 (+1.3) & 76.5 & 60.3 \\ 
\Xhline{2\arrayrulewidth}
\end{tabular}
}
\end{table*}

\subsubsection{LaSOT \cite{lasot}} is a recently published dataset that consists of 1,400 videos with more than 3.5M frames in total.
This benchmark is widely used to measure the long-term capability of trackers. 
The average video length of LaSOT is more than 2,500 frames, and each sequence comprises various challenging attributes.
The one-pass evaluation (OPE) protocol is used to measure the normalized precision ($\text{P}_{\text{Norm}}$) and the area under curve (AUC) of the success plot.
Table \ref{table:eval_base_slt} shows that the proposed method consistently improves all baseline trackers.

\subsubsection{TrackingNet~\cite{trackingnet}} is a large-scale dataset that provides 30K videos in training split and 511 videos in test split.
We evaluate our trackers on the test split of TrackingNet through the evaluation server.
Table \ref{table:eval_base_slt} shows that our SLT improves the AUC score by 7.6\%p, 2.6\%p, and 1.7\%p for SiamRPN++, SiamAttn, and TransT, respectively. It is noteworthy that the simple convolutional tracker SiamRPN++ shows competitive performance with SiamAttn with the power of SLT.

\subsubsection{GOT-10k~\cite{got10k}}
is a large-scale dataset that contains 10k sequences for training and 180 videos for testing.
For evaluation metrics, the average overlap (AO) and the success rate (SR) at overlap thresholds 0.5 and 0.75 are adopted.
Following the evaluation protocol of GOT-10k, we retrain our models using only the GOT-10k train split and submit the tracking results to the evaluation server.
Since the GOT-10k benchmark does not provide mask annotations, SLT-SiamAttn is trained without the mask branch.
Table \ref{table:eval_base_slt} shows that our SLT successfully improves all the baseline trackers in all evaluation metrics.
Baseline models are reproduced using only the GOT-10k train split.

\begin{table*}[t]
\centering
\caption{
Comparison with the state-of-the-art trackers on LaSOT.
}
\label{table:comparison_lasot}
\scalebox{0.75}{
\setlength{\tabcolsep}{2pt}
\begin{tabular}{c|cccccccccccc}
\Xhline{2\arrayrulewidth} 
\multirow{2}{*}{} & PACNet & Ocean & DiMP50 & PrDiMP50 & TransT & STARK- & STARK- & \textbf{SLT-} & \textbf{SLT-} & \textbf{SLT-} & \textbf{SLT-} \\
 & \cite{pacnet} & \cite{ocean} & \cite{dimp} & \cite{prdimp} & \cite{transt} & ST50 \cite{stark} & ST101 \cite{stark} & \textbf{SiamRPN++} & \textbf{SiamAttn} & \textbf{TrDiMP} & \textbf{TransT} \\ 
\hline
AUC (\%) & 55.3 & 56.0 & 56.9  & 59.8 & 64.2 & 66.4 & \textbf{67.1} & 58.4 & 57.4 & 64.4 & 66.8\\
$\text{P}_{\text{Norm}}$ (\%) & 62.8 & 65.1 & 64.3 & 68.0 & 73.7 & 76.3 & \textbf{77.0} & 66.6 & 66.2 & 73.5 & 75.5\\
\Xhline{2\arrayrulewidth}
\end{tabular}
}
\setlength{\abovecaptionskip}{10pt}
\setlength{\belowcaptionskip}{0pt}
\caption{
Comparison with the state-of-the-art trackers on TrackingNet.
}
\label{table:comparison_trackingnet}
\scalebox{0.73}{
\setlength{\tabcolsep}{2pt}
\begin{tabular}{c|cccccccccccc}
\Xhline{2\arrayrulewidth} 
\multirow{2}{*}{}  & DiMP50 & SiamFC++ & MAML & PrDiMP50 & TransT & STARK- & STARK- & \textbf{SLT-} & \textbf{SLT-} & \textbf{SLT-} & \textbf{SLT-}\\
 & \cite{dimp} & \cite{siamfcpp} & \cite{mamltrack} & \cite{prdimp}  & \cite{transt} & ST50 \cite{stark} & ST101 \cite{stark} & \textbf{SiamRPN++} &  \textbf{SiamAttn} & \textbf{TrDiMP} & \textbf{TransT}\\ 
\hline
AUC (\%) & 74.0 & 75.4 & 75.7 & 75.8 & 81.1  & 81.3 & 82.0 & 75.8 & 76.9 & 78.1 & \textbf{82.8} \\
$\text{P}_{\text{Norm}}$ (\%) & 80.1 & 80.0 & 82.2 & 81.6 & 86.8 & 86.1 & 86.9 & 81.0 & 82.3 & 83.1 &\textbf{87.5}\\
\Xhline{2\arrayrulewidth}
\end{tabular}
}
%
\centering
\caption{
Comparison with the state-of-the-art trackers on GOT-10k. `Add. data' denotes that trackers are trained using additional training datasets other than GOT-10k.
}
\label{table:comparison_got10k}
\scalebox{0.71}{
\setlength{\tabcolsep}{2pt}
\begin{tabular}{c|c|ccccccccccc}
\Xhline{2\arrayrulewidth} 
\multirow{2}{*}{} & Add. & SiamFC++ & DiMP50 & Ocean & PrDiMP50 & TransT & TrDiMP & STARK- & \textbf{SLT-} & \textbf{SLT-} & \textbf{SLT-} & \textbf{SLT-}  \\
& data & \cite{siamfcpp} & \cite{dimp} & \cite{ocean} & \cite{prdimp} & \cite{transt} & \cite{trdimp} & ST50 \cite{stark} & \textbf{SiamRPN++} & \textbf{SiamAttn} & \textbf{TrDiMP} & \textbf{TransT} \\ 
\hline
AO (\%) & \multirow{3}{*}{-} & 59.5 & 61.1 & 61.1 & 63.4 & 66.2 & 67.1 & \textbf{68.0} & 62.1 & 62.5 & 67.5 & 67.5  \\
$\text{SR}_{0.5}$ (\%) &  & 69.5 & 71.7 & 72.1 & 73.8 & 75.5 & 77.4 & 77.7 & 74.9 & 75.4 & \textbf{78.8} & 76.5  \\
$\text{SR}_{0.75}$ (\%) &  & 47.9 & 49.2 & 47.3 & 54.3 & 58.7 & 58.5 & \textbf{62.3} & 49.0 & 50.2 & 58.7 & 60.3 \\ \hline
AO (\%) & \checkmark  & - & 60.4 & - & 65.2 & 71.9 & 68.6 & 71.5 & 56.9 & 62.8 & 69.0 & \textbf{72.5}  \\
\Xhline{2\arrayrulewidth}
\end{tabular}
}
\end{table*}

\subsubsection{Comparison with SOTA trackers.}
We compare the performance of the proposed SLT family with the other state-of-the-art trackers on LaSOT and TrackingNet as shown in Table~\ref{table:comparison_lasot} and \ref{table:comparison_trackingnet}.
When compared to the recently proposed RL-based tracker PACNet~\cite{pacnet}, all four SLT trackers are showing superior performance by a large margin.
SLT-TransT, which is our best model, achieves state-of-the-art performance in both benchmarks.
Note that STARK-ST101~\cite{stark} uses deeper backbone (ResNet101) than TransT, which use ResNet50 backbone.
SLT-TransT thus needs to be compared with STARK-ST50 for fairness.
Table~\ref{table:comparison_got10k} also shows that both SLT-TrDiMP and SLT-TransT achieve comparable performance with state-of-the-art trackers on GOT-10k.

\subsection{Analysis}\label{sec:experiment_analysis}
We also analyze the effects of SLT using SiamRPN++ as the base tracker. The experimental analyses in this subsection are done on the \textit{validation} split of GOT-10k and the \textit{test} splits of LaSOT and TrackingNet.

\subsubsection{Sequence-level training components.}
\begin{figure}[t]
\centering
\includegraphics[width=1.0\linewidth]{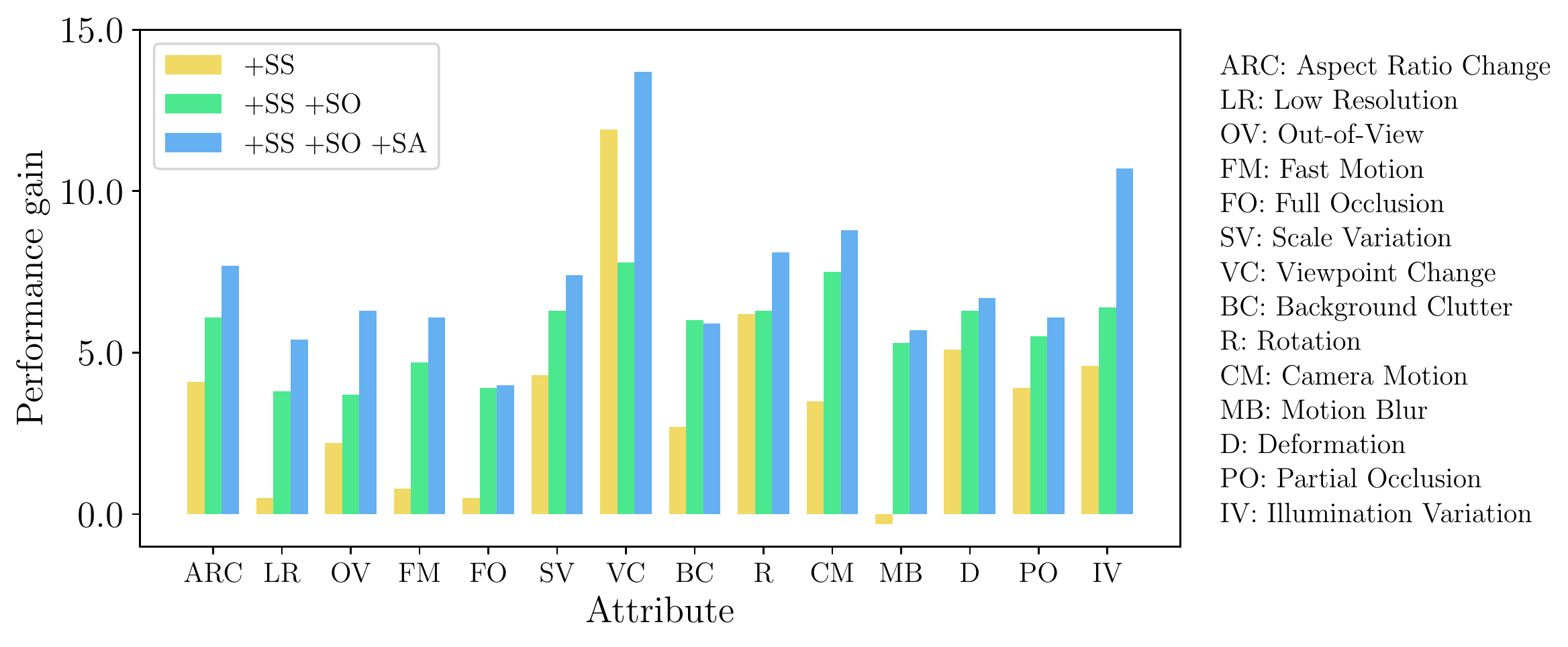}
\caption{
Benefits of sequence-level training components to individual attributes on the LaSOT dataset.
The baseline tracker is SiamRPN++, and the y-axis is performance (AUC) gain compared with the baseline tracker.
}
\label{fig:small_attr}
\end{figure}

The benefit of SLT comes from sequence-level sampling, sequence-level objective, and sequence-level data augmentation.
We validate the components of SLT by measuring the accuracy gains on the three benchmarks (Table~\ref{table:ssvsso}) and performing an attribute-based analysis on LaSOT (Fig.~\ref{fig:small_attr}).

\textit{Sequence-level sampling (SS).}
To measure the net effect of SS, we train a tracker with SS but use the frame-level objective.
As shown at `+SS' in Table~\ref{table:ssvsso}, by learning more accurate input data distribution, the tracker with SS outperforms the baseline with frame-level sampling by 4.1\%p, 5.3\%p, and 3.8\%p, respectively, on the three benchmarks. SS allows the tracker to observe realistic appearance variations of target objects during tracking, making itself more robust to variations of aspect ratio, scale, rotation, and illumination (ARC, SC, R, IV) as seen in Fig.~\ref{fig:small_attr}.

\begin{table*}[t]
\centering
\caption{
Effect of sequence-level training components.
}
\label{table:ssvsso}
\scalebox{0.85}{
\setlength{\tabcolsep}{5pt}
\begin{tabular}{c|cccc}
\Xhline{2\arrayrulewidth}
\multirow{2}{*}{Benchmark} & \multicolumn{4}{c}{SiamRPN++} \\  
 & Baseline & +SS ($\Delta$) & +SS+SO ($\Delta$) & +SS+SO+SA ($\Delta$) \\ \hline
 LaSOT (AUC) & 51.0 & 55.1 (+4.1)& 57.3 (+6.3) & 58.4 (+7.4) \\
 TrackingNet (AUC) & 68.2 & 73.5 (+5.3) & 75.0 (+6.8) & 75.8 (+7.6) \\
 GOT-10k (AO)   & 66.4 & 70.2 (+3.8) & 73.8 (+7.4) & 74.3 (+7.9) \\
\Xhline{2\arrayrulewidth}
\end{tabular}
}
\end{table*}

\textit{Sequence-level objective (SO).}
As shown at `+SS+SO' in Table~\ref{table:ssvsso}, SO additionally improves the performance by 2.2\%p, 1.5\%p, and 3.6\%p on three benchmarks, respectively.
SO enables the tracker to reflect accumulated localization errors, preventing it from losing the target in challenging situations such as full occlusion, background clutters, and motion blur (FO, BC, MB) as seen in Fig.~\ref{fig:small_attr}.

\textit{Sequence-level augmentation (SA).}
As shown at `+SS+SO+SA' in Table~\ref{table:ssvsso}, SA further improves the performance by 1.1\%p, 0.8\%p, and 0.5\%p, respectively, resulting in the significant gain of SLT in total as 7.4\%p, 7.6\%p, and 7.9\%p. The effectiveness of SA is also evident from the improvement in the overall attributes in Fig.~\ref{fig:small_attr}.

To show that the frame-interval augmentation strategy of SA also potentially helps in adapting to videos with diverse frame rates, we set up tracking scenarios with lower frame rates (\ie, faster motion).
In the evaluation protocol, we track objects  every $i$th frame only, skipping all the other frames; 
when the interval is 1, the evaluation protocol is the same as the original benchmark.
Table~\ref{table:frame_interval} shows that the frame-interval augmentation strategy not only improves the performance in normal videos, but also makes the tracker more robust to videos with lower frame rates.

\begin{table*}[t]
\centering
\caption{
Effect of sequence-level augmentation (SA) in terms of video frame interval with the low frame rate protocol. The frame interval is denoted by $i$.
}
\label{table:frame_interval}
\scalebox{0.85}{
\setlength{\tabcolsep}{6pt}
\begin{tabular}{c|c|ccc|cccc}
\Xhline{2\arrayrulewidth}
\multirow{2}{*}{Method} & \multirow{2}{*}{SA} & \multicolumn{3}{c|}{GOT-10k (AO)} & \multicolumn{4}{c}{LaSOT (AUC)} \\  
   &         & $i=1$ &$i=2$ &$i=3$& $i=1$ & $i=2$ & $i=3$ & $i=4$  \\ \hline
SiamRPN++       & -          & 66.4 & 63.1 & 60.8 & 51.0 & 50.0 & 50.2 & 48.8\\
SLT-SiamRPN++   & -          & 73.8 & 67.9 & 65.5 & 57.3 & 55.1 & 54.1 & 52.6\\
SLT-SiamRPN++   & \checkmark & 74.3 & 70.8 & 67.8 & 58.4 & 56.9 & 56.2 & 54.6\\
\Xhline{2\arrayrulewidth}
\end{tabular}
}
\setlength{\abovecaptionskip}{10pt}
\setlength{\belowcaptionskip}{0pt}
\centering
\caption{
Effect of training sequence length.
}
\label{table:seq_len}
\scalebox{0.85}{
\setlength{\tabcolsep}{8pt}
\begin{tabular}{c|cccccc}
\Xhline{2\arrayrulewidth}
\multirow{2}{*}{Benchmark} & \multicolumn{6}{c}{Training sequence length ($T$)} \\ 
 & 1 & 4 & 8 & 16 & 24 & 32 \\ \hline
GOT-10k (AO)   & 65.8 & 69.6 & 70.1 & 73.0 & \textbf{73.8} & 73.4\\
\Xhline{2\arrayrulewidth}
\end{tabular}
}

\centering
\caption{
Effect of frame-level pre-training. The zero (0) epoch stands for random initialization.
}
\label{table:pretrain_flt}
\scalebox{0.85}{
\setlength{\tabcolsep}{8pt}
\begin{tabular}{c|ccccc}
\Xhline{2\arrayrulewidth}
\multirow{2}{*}{Method} & \multicolumn{5}{c}{Frame-level pre-training (epoch)} \\ 
 & 0 & 1 & 5 & 10 & 20  \\ \hline
SiamRPN++       & - & 62.0 & 64.2 & 64.9 & 66.4\\ 
SLT-SiamRPN++   & 60.3 & 68.3 & 70.6 & 72.1 & 74.3\\ 
\Xhline{2\arrayrulewidth}
\end{tabular}
}
\end{table*}

\subsubsection{Length of training episodes.}
In training time, we randomly sample training episodes of pre-defined sequence length $T$.
Learning temporal dependency may be affected by the length of training sequences.
We thus experiment with varying $T$ while fixing the sampled frame interval to 1.
The best result is obtained with $T=24$, as can be seen in Table~\ref{table:seq_len}.
When $T=1$, the performance does not improve over the pre-trained tracker. 

\subsubsection{Frame-level pre-training.}
In our experiments, we perform SLT from a model pre-trained by frame-level training (FLT).
We analyze the effect of  warm-up FLT in Table~\ref{table:pretrain_flt}, showing that it improves tracking performance indeed, and FLT with only a few epochs is sufficient for the warm-up.
We also observed that the gain from SLT easily disappears by another few epochs of FLT (\ie, FLT $\to$ SLT $\to$ FLT).
In particular, the AO score of SiamRPN++ on the GOT-10k validation split reverted from 74.3\% to 66.2\% after 5 epochs of FLT.
This indicates that SLT grants a unique gain, which FLT cannot provide.

$\\$
For additional analysis and qualitative results, see the supplementary material.

%% file: _5_conclusion.tex

\section{Conclusion}
We have proposed a novel sequence-level training strategy for visual object tracking to resolve the training-testing inconsistency problem of existing trackers.
Unlike existing methods, it trains a tracker by actually tracking on a video and directly optimizing a tracking performance metric,   
 boosting the generalization performance without modifying the model architecture.
Experiments on SiamRPN++, SiamAttn, TransT, and TrDiMP trackers show that sequence-level sampling, objective, and augmentation are all effective in learning visual tracking. 

%% file: _6_supple.tex


\title{Supplementary Material for \\Towards Sequence-Level Training for Visual Tracking} 

\titlerunning{Towards Sequence-Level Training for Visual Tracking}
%
\author{Minji Kim\textsuperscript{1*} \quad
Seungkwan Lee\textsuperscript{3,4*}\quad
Jungseul Ok\textsuperscript{3} \quad
Bohyung Han\textsuperscript{1,2} \quad
Minsu Cho\textsuperscript{3}
}

\authorrunning{M. Kim et al.}
%

\institute{\textsuperscript{1}ECE \& \textsuperscript{2}IPAI, Seoul National University\\
\textsuperscript{3}Pohang University of Science and Technology (POSTECH)\\
\textsuperscript{4}Deeping Source Inc. \\
\url{https://github.com/byminji/SLTtrack}
}

\maketitle
\def\thefootnote{}\footnotetext{* These authors contributed equally to this work.}\def\thefootnote{\arabic{footnote}}

\appendix
\setcounter{table}{0}
\definecolor{new_yellow}{rgb}{1.0, 0.85, 0.0}

\section{Training Details}
We use Adam optimizer for SiamRPN++, SiamAttn, and TrDiMP, and AdamW optimizer for TransT.
SiamRPN++ and SiamAttn are trained for 20 epochs with 10000 videos per epoch, while the learning rate starts from $10^{-5}$ and exponentially decays to $10^{-6}$.
Following the original papers, we use $0.1$ times smaller learning rate for backbone layers for SiamRPN++ and $0.05$ times smaller for SiamAttn, respectively.
TransT is trained for 120 epochs with 1000 videos per epoch, and the learning rate starts from $0.1$ times the original model setup in the paper and decreases by a factor of 10 after 100 epochs.
TrDiMP is trained for 40 epochs with 5000 videos per epoch, and the learning rate starts from $0.04$ times the initial learning rate from the original model and halves every 8 epochs.
After the pre-training stage, the statistics of batch normalization layers are fixed and not updated during the RL fine-tuning stage.

\section{Evaluation on Additional Benchmarks}

\begin{table*}[ht]
\vspace{-9mm}
\centering
\caption{
Experimental results on NFS, UAV123, TNL2K, VOT2018, and VOT2020 baseline analysis.
Test-time hyper-parameters are tuned only in VOT.
}
\vspace{0.1cm}
\label{table:additional_benchmarks}
\scalebox{0.9}{
\hspace{-2mm}
\setlength{\tabcolsep}{6pt}
\begin{tabular}{cc|ccc|ccc|ccc}
\Xhline{2\arrayrulewidth}
\multicolumn{2}{c|}{\multirow{2}{*}{Method}} & NFS & UAV123 & TNL2K & \multicolumn{3}{c|}{VOT2018} & \multicolumn{3}{c}{VOT2020}\\  
   &        & AUC & AUC & AUC & EAO & AO* & AUC* & EAO & A & R\\ \hline
\multirow{2}{*}{SiamRPN++} & Base & 50.5 & 59.3 & 38.8 & \textbf{39.7} & 45.2  & 44.9 & 24.3 & \textbf{45.5} & 65.5\\
                           & +SLT & \textbf{56.5} & \textbf{61.2} & \textbf{44.1} & 34.3 & \textbf{48.8} & \textbf{48.6} & \textbf{25.4} & 45.3 & \textbf{70.8}\\ 
                          \hline
\multirow{2}{*}{TrDiMP}    & Base & 65.0 & 64.8 & 49.8 & 44.8 & 53.6 & 53.2 & 28.2 & 46.8 & 74.7\\
                           & +SLT & \textbf{65.6} & \textbf{66.3} & \textbf{50.7} & \textbf{45.0} & \textbf{54.1} & \textbf{53.6} & \textbf{28.9} & \textbf{47.0} & \textbf{76.2} \\ 
                          \hline
\multirow{2}{*}{TransT}    & Base & 65.3 & 66.6 & 53.5 & 30.0 & 51.0 & 50.6 & \textbf{29.3} & \textbf{47.7} & 75.3\\
                           & +SLT & \textbf{66.2} & \textbf{68.6} & \textbf{55.0}  & \textbf{30.6} & \textbf{52.6} & \textbf{52.3} & \textbf{29.3} & 46.7 & \textbf{76.0} \\ 
\Xhline{2\arrayrulewidth}
\end{tabular}
}
\end{table*}

\vspace{-3mm}
\noindent
Table~\ref{table:additional_benchmarks} summarizes the evaluation results of our method on additional benchmarks, where SLT consistently improves the AUC scores on NFS~\cite{nfs}, UAV123~\cite{uav}, and TNL2K~\cite{tnl2k}, while strengthening the robustness (R) on VOT2020~\cite{vot2020}.
Note that the re-initialization policy of VOT evaluation does not match with the reward system of SLT, which is designed for one-pass evaluation. Therefore, we also compare AO and AUC on VOT2018~\cite{vot2018}, showing that SLT further benefits when the test-time metric and the reward system are aligned.
Because VOT2020 does not provide the bounding box annotation, we cannot report AO and AUC.

\section{Qualitative Results}
\begin{figure*}[ht]
\centering
\begin{subfigure}{0.95\textwidth}
    \includegraphics[width=\textwidth]{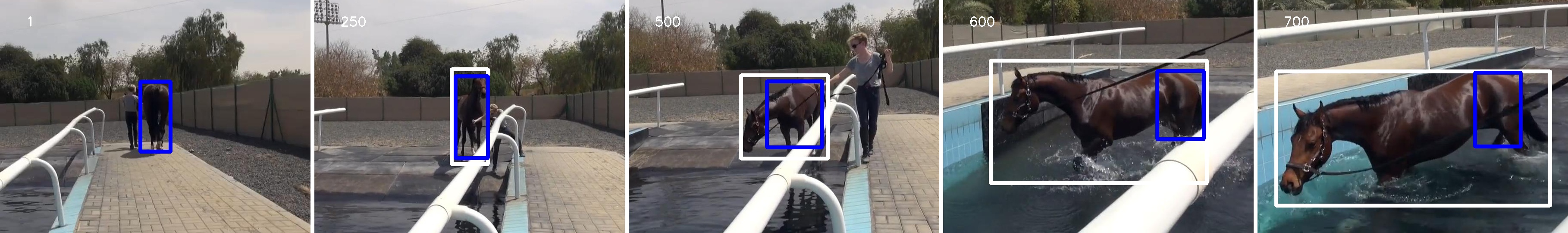}
    \includegraphics[width=\textwidth]{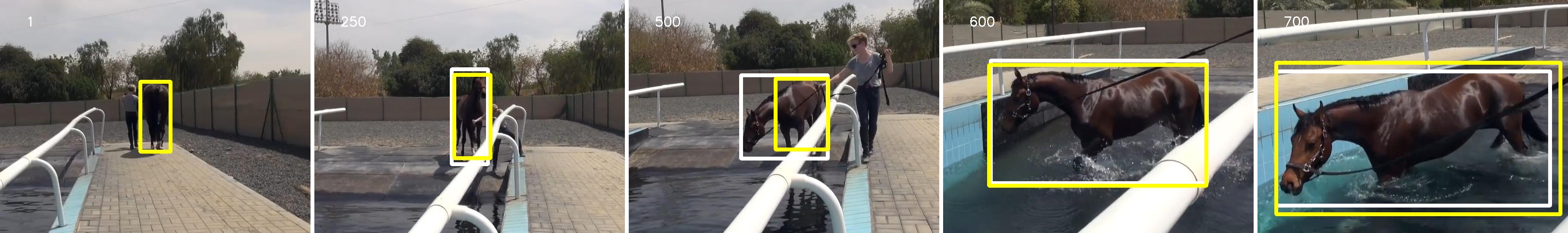}
    \includegraphics[width=\textwidth]{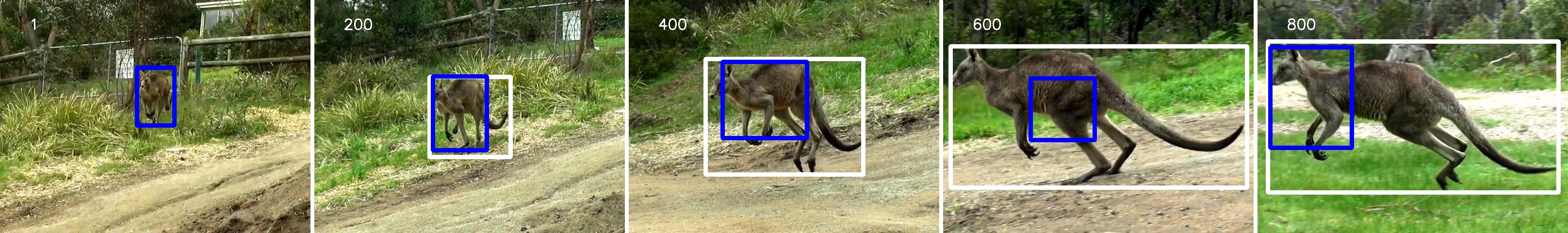}
    \includegraphics[width=\textwidth]{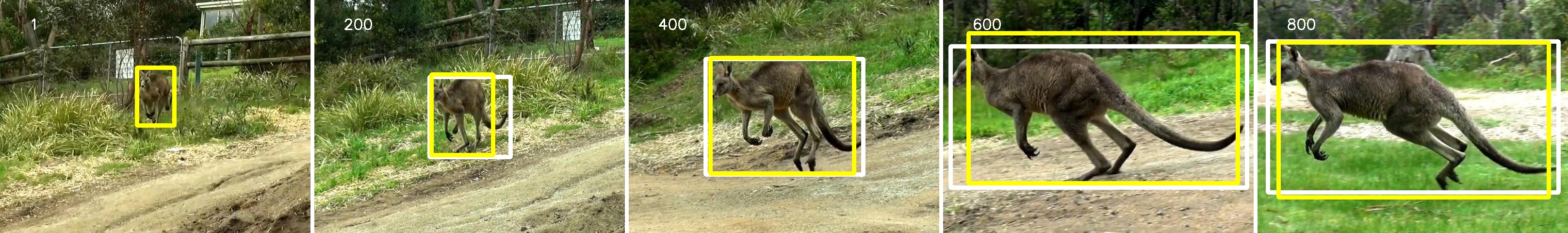}
    \caption{{\color{blue}Baseline} \textit{vs.} {\color{new_yellow}Baseline+SS}}
    \vspace{0.5cm}
    \label{fig:basevsss}
\end{subfigure}
\begin{subfigure}{0.95\textwidth}
    \includegraphics[width=\textwidth]{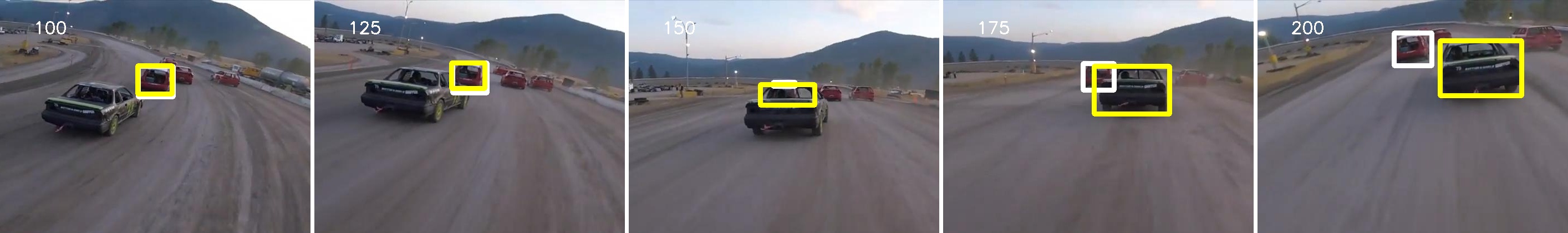}
    \includegraphics[width=\textwidth]{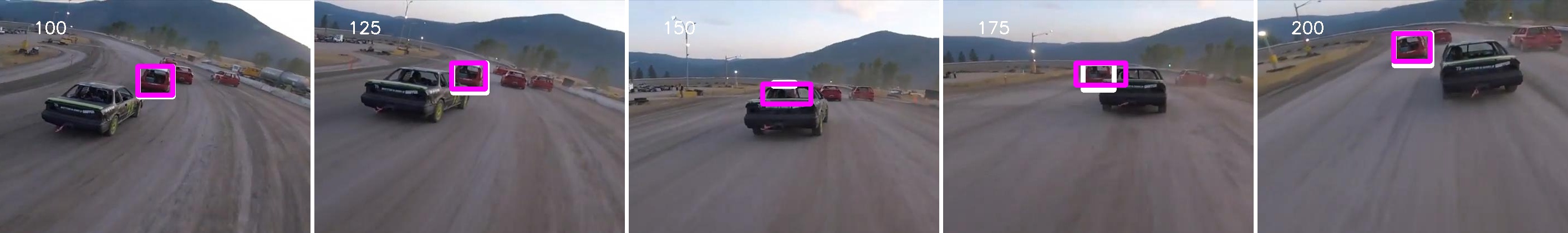}
    \includegraphics[width=\textwidth]{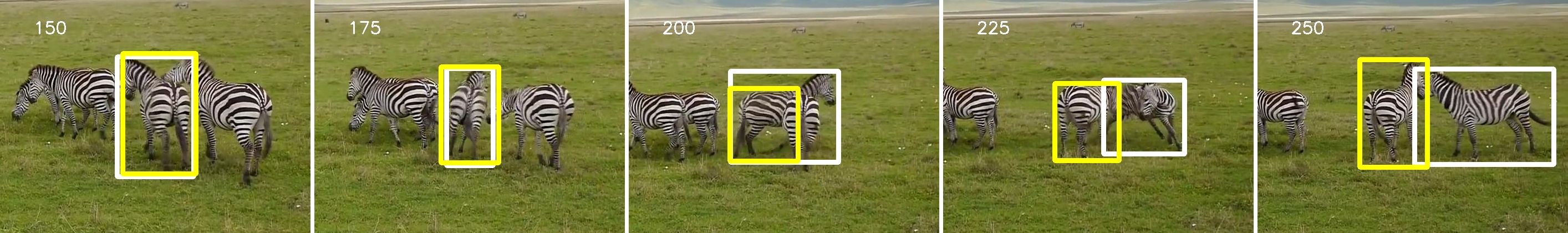}
    \includegraphics[width=\textwidth]{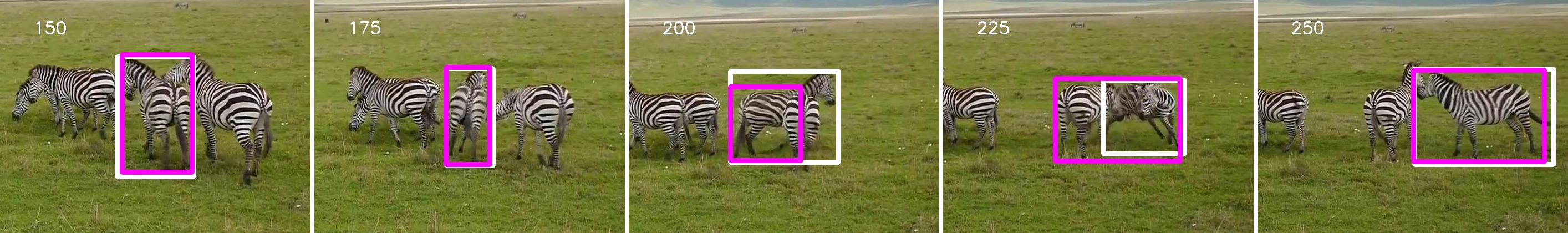}
    \caption{{\color{new_yellow}Baseline+SS} \textit{vs.} {\color{magenta}Baseline+SS+SO}}
    \label{fig:ssvsso}
\end{subfigure}
\caption{
Qualitative results.
white: ground truth, {\color{blue}blue}: baseline, {\color{new_yellow}yellow}: tracker trained with SS, {\color{magenta}magenta}: tracker trained with SS and SO.
}
\label{fig:qual}
\end{figure*}

Section 4.4 of the main paper describes the benefits of sequence-level sampling (SS) and sequence-level objective (SO).
To support our argument about the effect of these two components, we present qualitative results in Fig.~\ref{fig:qual}, which visualizes the bounding-boxes corresponding to ground truth (white) and results from {\color{blue} baseline (blue)}, {\color{new_yellow}baseline+SS (yellow)}, and {\color{magenta}baseline+SS+SO (magenta)}.
For this analysis, we adopt SiamRPN++~\cite{siamrpnpp} as the baseline tracker.
As discussed in the main paper, SS makes trackers more robust to appearance updates given by scale changes, aspect ratio variations, and rotation. 
In Fig.~\ref{fig:basevsss}, there are two videos whose target objects change their appearance significantly.
The tracker trained with SS successfully adapts to appearance variations, while the baseline tracker fails to capture the entire bodies of the target objects.
Moreover, SO alleviates the drift issue in trackers in some challenging situations such as occlusion and background clutter.
To qualitatively validate the properties, we present the target trajectories of the {\color{new_yellow}baseline+SS} and {\color{magenta}baseline+SS+SO} trackers in Fig.~\ref{fig:ssvsso}, which includes the videos with such challenging attributes.

